\documentclass{article}

\usepackage{arxiv}

\usepackage[utf8]{inputenc} % allow utf-8 input
\usepackage[T1]{fontenc}    % use 8-bit T1 fonts
\usepackage{hyperref}       % hyperlinks
\usepackage{url}            % simple URL typesetting
\usepackage{booktabs}       % professional-quality tables
\usepackage{amsfonts}       % blackboard math symbols
\usepackage{nicefrac}       % compact symbols for 1/2, etc.
\usepackage{microtype}      % microtypography
\usepackage{lipsum}
\usepackage{graphicx}
\usepackage{amsmath}
\usepackage{algorithm,algorithmicx,algpseudocode}
\usepackage{amsthm}
\usepackage{subcaption}
\usepackage{mathtools}
\usepackage{varwidth}
\usepackage{enumitem}
\usepackage{wrapfig}

\usepackage{footnote}
\makesavenoteenv{tabular}
\makesavenoteenv{table}

\newcommand{\precap}{\vspace{-0pt}}
\newcommand{\postcap}{\vspace{-0pt}}

\newcommand{\presec}{\vspace{-0pt}}
\newcommand{\postsec}{\vspace{-0pt}}
 
\title{LightMC: A Dynamic and Efficient Multiclass Decomposition Algorithm}

\author{
  Ziyu Liu\thanks{The work is done when the author was visiting Microsoft Research.} \\
  Peking University \\
  \texttt{ziyu-liu@pku.edu.cn} \\
  \And
  Guolin Ke \\
  Microsoft Research \\
  \texttt{guolin.ke@microsoft.com} \\
  \AND
  Jiang Bian \\
  Microsoft Research \\
  \texttt{jiang.bian@microsoft.com} \\
  \And
  Tie-Yan Liu \\
  Microsoft Research \\
  \texttt{tyliu@microsoft.com} \\
}  

\begin{document}
\maketitle

\begin{abstract}
Multiclass decomposition splits a multiclass classification problem into a series of independent binary learners and recomposes them by combining their outputs to reconstruct the multiclass classification results. Three widely-used realizations of such decomposition methods are One-Versus-All (OVA), One-Versus-One (OVO), and Error-Correcting-Output-Code (ECOC). While OVA and OVO are quite simple, both of them assume all classes are orthogonal and neglect the latent correlation between classes in real-world. Error-Correcting-Output-Code (ECOC) based decomposition methods, on the other hand, are theoretically preferable due to its integration of the correlation among classes. However, the performance of existing ECOC-based methods highly depends on the design of coding matrix and decoding strategy. Unfortunately, it is quite uncertain and time-consuming to discover an effective coding matrix with appropriate decoding strategy. To address this problem, we propose LightMC, an efficient dynamic multiclass decomposition algorithm. Instead of using fixed coding matrix and decoding strategy, LightMC uses a differentiable decoding strategy, which enables it to dynamically optimize the coding matrix and decoding strategy, toward increasing the overall accuracy of multiclass classification, via back propagation jointly with the training of base learners in an iterative way. Empirical experimental results on several public large-scale multiclass classification datasets have demonstrated the effectiveness of LightMC in terms of both good accuracy and high efficiency.
\end{abstract}

\presec
\section{Introduction}\label{intro}
\postsec

Multiclass classification is the problem of classifying data instances into one of three or more classes. In a typical learning process of multiclass classification, assuming that there are $K > 2$ classes, i.e. $Y= \{C_1,C_2,...,C_K\}$, and $n$ training instances, i.e. $S=\{\{x_1,y_1\},\{x_2,y_2\}...,\{x_n,y_n\}\}$, each training instance belongs to one of $K$ different classes, and the goal is to construct a function $f(x)$ which, given a new data instance $x$, can correctly predict the class to which the new instance belongs. Multiclass classification problems are very common in the real-world with a variety of scenarios, such as image classification~\cite{ciregan2012multi}, text classification~\cite{nigam2000text}, e-commerce product classification~\cite{schulten2001commerce}, medical diagnosis~\cite{panca2017application}, etc. Currently, one of the most widely-used solutions for multiclass classification is the decomposition methods\footnote{There are also some other efforts that trying to solve multiclass problem directly, like~\cite{bredensteiner1999multicategory,choromanska2015logarithmic,mroueh2012multiclass,weston1998multi, hsu2009multi,prabhu2014fastxml,si2017gradient,yen2016pd}. However, they are not as popular as decomposition methods and thus are not in the scope of this paper.}, which splits a multiclass problem, or polychotomy, into a series of independent two-class problems (dichotomies) and recompose them using the outputs of dichotomies in order to reconstruct the original polychotomy. In practice, the widespread use of decomposition methods is mainly due to its simplicity and easy-adaptation to existing popular learners, e.g. support vector machines, neural networks, gradient boosting trees, etc.

There are a couple of concrete realization of decomposition methods, including One-Versus-All (OVA)~\cite{nilsson1965learning}, One-Versus-One (OVO)~\cite{hastie1998classification}, and Error-Correcting-Output-Code (ECOC)~\cite{dietterich1995solving}.In particular, OVA trains $K$ different base learners, for the $i$-th of which let the positive examples be all the instances in class $C_i$ and the negative examples be all not in $C_i$; OVO trains $K\times(K-1) / 2$ base learners, one of each to distinguish each pair of classes. While OVA and OVO are simple to implement and widely-used in practice, they yield some obvious disadvantages. First, both OVA and OVO are based on the assumption that all classes are orthogonal and the corresponding base learners are independent with each other, which, nevertheless, neglect the latent correlation between these classes in real-world applications. For example, in a task of image classification, the instances under the `{\em Cat}' class apparently yield stronger correlation to those under the `{\em Kitty}' class than those under the `{\em Dog}' class. Moreover, the training of OVA and OVA is inefficient since its high computation complexity when $K$ is large, leading to extremely high training cost when processing large-scale classification datasets.

ECOC-based methods, on the other hand, are theoretically preferable over both OVA and OVO since it can in some sense alleviate their disadvantages. More concretely, ECOC-based methods rely on a coding matrix, which defines a new transformation of instance labeling, to decompose the multiclass problem into dichotomies, and then recompose in a way that makes decorrelations and correct errors. Generating different distances for different pairs of classes, indeed, enable ECOC-based methods to leverage the correlations among classes into the whole learning process. For example, if the coding matrix assigns $(1, 1, 1)$, $(1, 1, -1)$ and $(-1, -1, -1)$ to `{\em Cat}', `{\em Kitty}' and `{\em Dog}', respectively, the learned model can ensure a closer distance between instance pairs across `{\em Cat}' and `{\em Kitty}' than those across `{\em Cat}' and `{\em Dog}'. Moreover, since the length of the code, also the number of base learners, could be much smaller than $K$, ECOC-based methods can significantly reduce the computation complexity over OVA and OVO, especially when the original class number $K$ is very large. 

Given the delicate design of class coding, the performance of ECOC-based methods highly depends on the design of the coding matrix and the corresponding decoding strategy. The most straightforward way is to create a random coding matrix for class transformation with Hamming decoding strategy. The accuracy of this simple approach, apparently, can be of highly volatile due to its randomness. To address this problem, many efforts have been made focusing on optimizing the coding matrix. However, it is almost impossible to find an optimal coding matrix due to its complexity and even finding a sub-optimal coding matrix is likely to be quite time-consuming. Such uncertainty and inefficiency in recognizing a sub-optimal coding matrix undoubtedly prevent the broader using of the ECOC-based methods in real-world scenarios. 

To address this challenge, we propose a new dynamic ECOC-based decomposition approach, named LightMC. Instead of using fixed coding matrix and decoding strategy, LightMC can dynamically optimize the coding matrix and decoding strategy, toward more accurate multiclass classification, jointly with the training of base learners in an iterative way. 
To achieve this, LightMC takes advantage of a differentiable decoding strategy which allows it to perform the optimization by gradient descent, guarantees that the training loss can be further reduced.
In addition to improving final classification accuracy and obtaining the coding matrix and decoding strategy more beneficial to the classification performance, LightMC can, furthermore, significantly boost the efficiency since it saves much time for searching sub-optimal coding matrix. As LightMC will optimize coding matrix together with the model training process, it is not necessary to spend much time in tuning an initial coding matrix, and, as shown by further empirical studies, even a random coding matrix can result in satisfying.  

To validate the effectiveness and efficiency of LightMC, we conduct experimental analysis on several public large-scale datasets. The results illustrate that LightMC can outperform OVA and existing ECOC-based solution on both training speed and accuracy. 

This paper has following major contributions:

\begin{itemize}[leftmargin=12pt,itemsep=1pt,topsep=2pt]
\item We propose a new dynamic decomposition algorithm, named LightMC, that can outperform traditional ECOC-based methods in terms of both accuracy and efficiency.
\item We define a differentiable decoding strategy and derive an effective algorithm to dynamically refine the coding matrix by extending the well-known back propagation algorithm.
\item Extensive experimental analysis on multiple public large-scale datasets to demonstrate both the effectiveness and the efficiency of proposed new decomposition algorithm is highly efficient.
\end{itemize}

The rest of the paper is organized as followed. Section~\ref{sec_background} introduces ECOC decomposition approaches and related work. Section~\ref{lightmc} presents the details of the LightMC. Section~\ref{experiment} shows experiment results that validate our proposition on large-scale public available multiclass classification data sets. Finally, we conclude the paper in Section~\ref{conclusion}.

\presec
\section{Preliminaries}\label{sec_background}
\postsec

\subsection{Error Correcting Output Code (ECOC)} \label{sec_ecoc}
\postsec
ECOC was first introduced to decompose multiclass classification problems by
Dietterich and Bakiri~\cite{dietterich1995solving}. In this method, each class $k$ is assigned to a codeword $\boldsymbol{M_k}$, where $M_{kj}$ represents the label of data from class $k$ when learning the base learner $j$. All codewords can be combined to form a matrix $\boldsymbol{M}\in \{-1,1\}^{K\times L}$, where $L$ is the length of one codeword as well as the number of base learners. Given the output of base learners $\boldsymbol{o} = \{o_1, o_2,\dots,o_L\}$, the final multiclass classification result can be obtained through a decoding strategy:
\begin{equation}
\label{original_decoding}
    \hat{\boldsymbol{y}}= argmin_k(\boldsymbol{t}), \textrm{\ \ where\ \ } t_k = \frac{1}{2}\sum_{j=1}^{L}{\vert M_{kj}-sgn(o_j) \vert}
\end{equation}

where $\hat{\boldsymbol{y}}$ is the predicted class and $sgn$ is the sign function and $sgn(o)$ equals 1 if $o \geq 0$ otherwise $-1$. This decoding strategy is also called hamming decoding as it makes the prediction by choosing the class with lowest hamming distance. Under such decoding strategy, the coding matrix is capable of correcting a certain amount of errors made by base learners~\cite{dietterich1995solving}. 

ECOC-based methods yield many advantages over traditional decomposition approaches. First, the introducing of the coding matrix, which can indicate different distances between different class pairs, indeed enables us to integrate the correlation among classes into the classification modeling so as to further improve the classification accuracy. Moreover, since code length $L$, i.e., the number of base learners, could be much smaller than the number of classes $K$, ECOC-based methods can be more efficient than OVA and OVO, especially when $K$ is very large. 

It is obvious that the classification performance of ECOC-based methods highly depend on the design of coding matrix. Nevertheless, the complexity of finding the best coding matrix is NP-Complete as stated in~\cite{crammer2002learnability}. Thus, it is almost impossible to find an optimal coding matrix, and even finding a sub-optimal coding matrix is likely to be quite time-consuming. Such uncertainty and inefficiency in finding a sub-optimal coding matrix undoubtedly prevent the broader using of the ECOC-based methods in real-world applications. 

\presec
\subsection{Related work}\label{relatedwork}
\postsec

Recent years have witnessed many efforts attempting to improve ECOC-based decomposition methods. Especially, many of existing studies focused on discovering more appropriate coding matrix. For example, some efforts made hierarchical partition of the class space to generate corresponding code~\cite{baro2009traffic,pujol2006discriminant}; some other studies explored the genetic algorithm to produce coding matrix with good properties~\cite{garcia2008evolving, bautista2012minimal, bagheri2013genetic, bautista2014design}; moreover, there are a couple of efforts that have demonstrated significant improvement on ECOC-based methods by using spectral decomposition to find a good coding matrix~\cite{zhang2009spectral} or by relaxing the integer constraint on the coding matrix elements so as to adopting a continuous-valued coding matrix~\cite{zhao2013sparse}. In the meantime, some previous studies turned to optimizing the decoding strategy by employing the bagging and boosting approach~\cite{hatami2012thinned, rocha2014multiclass} or assigning deliberate weigmost of previoushts on base learners for further aggregation~\cite{escalera2006ecoc}. 

While these previous studies can improve ECOC-based methods in some sense, they still suffer from two main challenges: 
1) \emph{Efficiency:} In order to increase multiclass classification accuracy, many of previous works like~\cite{baro2009traffic,pujol2006discriminant} designed the coding matrix with a long code length $L$, ranging from $K-1$ to $K^2$, which leads to almost as many base learners as models needed in OVA and OVO. Such limitation makes existing ECOC-methods very inefficient in the large-scale classification problems.
2) \emph{Scalability:} In fact, most of the previous ECOC-based methods were studied under a small-scale classification data, which usually consists of, for example, tens of classes and thousands of samples~\cite{zhao2013sparse}. To the best of knowledge, there is no existing deep verification of the performance of ECOC-based methods on a large-scale classification data. Meanwhile, such investigation is even quite difficult theoretically, since most of them cannot scale up to the large-scale data due to the long coding length and expected great pre-processing cost.

Because of these major shortages, it is quite challenging in applying existing ECOC-based methods into the real-world applications, especially those large-scale multiclass classification problems.

\presec
\section{LightMC}\label{lightmc}
\postsec

To address those major shortages of ECOC-based methods stated in Sec.~\ref{sec_background}, we proposed a new multiclass decomposition algorithm, named LightMC. Instead of determining the coding matrix and decoding strategy before training, LightMC attempts to dynamically refine ECOC decomposition by directly optimizing the global objective function, jointly with the training of base learners. 
More specifically, LightMC introduces a new differentiable decoding strategy, which enables LightMC to optimize the coding matrix and decoding strategy directly via gradient descent during the training of base learners. As a result, LightMC yields two-fold advantages: 1) \emph{Effectiveness:} rather than separate the designing of coding matrix and decoding strategy from the base learning training, LightMC can further enhance ECOC-based methods in terms of classification accuracy by jointly optimizing the coding matrix, decoding strategy, and base learners; 2) \emph{Efficiency:} since the coding matrix will be automatically optimized in the subsequent training, LightMC can significantly reduce time cost for finding a good coding matrix before training, 

In this section, we will first introduce the overall training algorithm. Then, we will present our new decoding model and derive the optimization algorithms for decoding strategy and coding matrix based on it. Moreover, we will take further discussions on the performance and efficiency of LightMC.

\presec
\subsection{Overall Algorithm}\label{overall_algorithm}
\postsec

\begin{figure}[H]
    \centering
    \includegraphics[height=1.1cm, width=11.8cm]{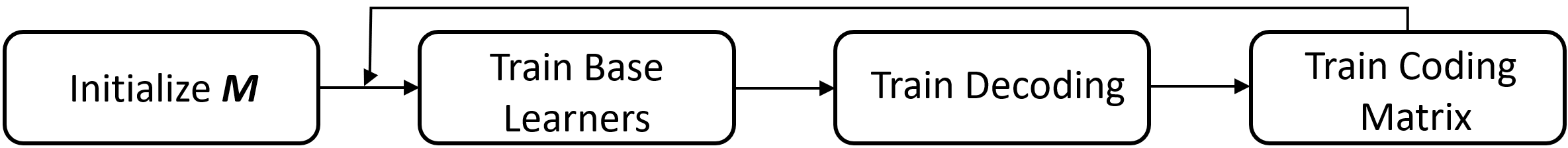}
    \caption{The general learning procedure of LightMC.}
    \label{flowchart}
\end{figure}

\begin{minipage}[t]{0.49\textwidth}
   \begin{algorithm}[H] 
   \caption{LightMC}
   \label{algo_train}
\begin{algorithmic}[1]
   \State {\bfseries Input:} data $\boldsymbol{X}$, label $\boldsymbol{y}$, code length $L$, \\base learner $f$, max iteration (epoch) times $T$, starting iteration $i_s$, base learner learning rate $\alpha$, decoding parameter $\boldsymbol{\Theta}$
   \State Initialize a coding matrix $\boldsymbol{M}$
   \State Initialize $\boldsymbol{\Theta}$ according to $\boldsymbol{M}$
   \For{$i=1$ {\bfseries to} $T$}
   \State Train base learner $f_1,f_2,...,f_L$ for a single iteration using $\boldsymbol{X},\boldsymbol{y},\boldsymbol{M}$ and $\alpha$
   \If{base learner is not boosting learner {\bfseries or} ($i\geq i_s$ {\bfseries and} ($i-i_s)\mod{1/\alpha} = 0$)}
   \State $\boldsymbol{o} = Predict(f_1, .., f_L)$
   \State $\hat{\boldsymbol{y}} = Decode(\boldsymbol{\Theta}, \boldsymbol{o})$ 
   \State {\bfseries TrainDecoding($\boldsymbol{\Theta}, \boldsymbol{y},\hat{\boldsymbol{y}}$)} 
   \State {\bfseries TrainCodingMatrix($\boldsymbol{M}, \boldsymbol{y},\hat{\boldsymbol{y}}$)}
   \EndIf
   \EndFor
\end{algorithmic}
\end{algorithm}
\end{minipage}
\hspace{1mm}
\begin{minipage}[t]{0.49\textwidth}
\begin{algorithm}[H]
   \caption{TrainDecoding}
   \label{algo_decoding}
\begin{algorithmic}[1]
   \State {\bfseries Input:} parameter $\boldsymbol{\Theta}$, label $\boldsymbol{y}$, prediction $\hat{\boldsymbol{y}}$
   \State Use mini-batch gradient descent to update $\boldsymbol{\Theta}$ 
\end{algorithmic}
\end{algorithm}
\begin{algorithm}[H]
   \caption{TrainCodingMatrix}
   \label{algo_update}
\begin{algorithmic}[1]
   \State {\bfseries Input:} coding matrix $\boldsymbol{M}$, label $\boldsymbol{y}$, prediction $\hat{\boldsymbol{y}}$
   \State Compute $\boldsymbol{G}$: data-wise gradients 
   \State Compute $\boldsymbol{C}$: \#data for each class
   \For{$i=1$ {\bfseries to} $L$} \Comment{For all codes}
   \State {Compute $\boldsymbol{S}$: sum gradients at $i$th code for each class} 
   \For{$k=1$ {\bfseries to} $K$} 
   \State $\beta_k=S_k/C_k$
   \State $M_{ki}=M_{ki}-\gamma_2\beta_k$
   \EndFor
   \EndFor
\end{algorithmic}
\end{algorithm}
\end{minipage}

The general learning procedure of LightMC is summarized as shown in~Fig.~\ref{flowchart}. More specifically, before LightMC starts training, a coding matrix is first initialized by existing ECOC-based solutions. Then, to make full use of training information from base learners, LightMC employs an alternating optimization algorithm, which alternates the learning of base learners together with the coding and decoding optimization: when training base learners, the coding and decoding strategy is fixed when training base learners, and vice versa. This joint learning procedure will run repeatedly until the whole training converges. 

Note that, instead of determining coding matrix before training, LightMC develops an end-to-end solution to jointly train base learners and the decomposition models in an iterative way. The details of the LightMC algorithm can be found in Alg.~\ref{algo_train}. Within this algorithm, there are two essential steps: \textbf{TrainDecoding} is used to optimize the decoding strategy, the details of which will be revealed in Sec.~\ref{sec_traindecoding}; and,  \textbf{TrainCodingMatrix} aims at optimizing the coding matrix, the details of which will be introduced in Sec.~\ref{codematrix_update}.

\presec
\vspace{-5pt}
\subsection{New Differentiable Decoding Strategy: Softmax Decoding}\label{sec_traindecoding}
\postsec

To find the optimal coding and decoding strategies, it is necessary to optimize directly on the global objective function. However, since most existing decoding strategies are not differentiable, it prevents us from optimizing the global objective function directly by employing widely-used back propagation method. To remove this obstacle, it is critical to design a decoding strategy which is differentiable while preserving error correcting properties.

A deep-dive into the decoding strategy, i.e., Eq.~\ref{original_decoding}, discloses two non-differentiable functions: $sgn$ and $argmin$. As introduced in~\cite{escalera2010decoding}, $sgn$ can be removed directly, since the resulting distance function will become Manhattan (L1) distance, which still preserves its error correcting property. In the meantime, $argmin$ can be replaced by the widely-used $softmax$, which is able to approximate $argmin$ with producing continuous probabilities and thus differentiable. More specifically, we can first replace the $argmin$ to $argmax$ by reversing the sign of $M_{kj}$ at the same time. In this way, when the output of the $j$-th classifier $o_j$ equals to $M_{kj}$, the distance will be the maximum value instead of the minimum. After that, we can replace the $argmax$ to $softmax$ directly, and the whole decoding strategy becomes
\begin{equation}
\label{equ_new_coding}
   \hat{\boldsymbol{y}}= softmax(\boldsymbol{t}), \textrm{\ \ where\ \ } t_k = \frac{1}{2}\sum_{j=1}^{L}{\vert -M_{kj}-o_j  \vert}, 
\end{equation}

where $t_k$ denotes the similarity between the classifier output and the code of class $k$.  Although the L1 loss is applied in the algorithm, L2 loss or other distance functions mentioned in~\cite{escalera2010decoding} are also applicable and should produce similar results. Note that, after all the transformation mentioned above, the decoding strategy will assign the highest score to the class closest to the output vector, which, in other words, is exactly the error-correcting property~\cite{escalera2010decoding}.

Recognizing such differentiable error correcting decoding strategy enables us to employ the widely-used gradient descent algorithm to optimize the decoding strategy directly. Before doing this, we notice that the new decoding function can be rewritten into a form of single layer softmax regression. As the distance function in Eq.~\ref{equ_new_coding} satisfies
\begin{equation*}
  \vert -M_{kj}-o_j \vert =
    \begin{cases}
      1-o_j, &M_{kj}=-1 \\
      1+o_j, &M_{kj}=1
    \end{cases}
    = 1 + M_{kj}o_j ,
\end{equation*}
it allows the decoding strategy to be rewritten into:
\begin{equation}
\label{equ_softmax}
\begin{split}
    t_k = \frac{1}{2}\sum_{j=1}^L{(1 + M_{kj}o_j)} &= \frac{1}{2}(\boldsymbol{M_k}^{T}\boldsymbol{o}+L), 
\textrm{\ \ let\ } \boldsymbol{\theta_k}=\boldsymbol{M_k}, b_k=L,\textrm{\ \ then we have}\\
    \boldsymbol{\hat{y}} &= softmax(\boldsymbol{t}), \ \  t_k = \frac{1}{2}(\boldsymbol{\theta_k}^T\boldsymbol{o}+b_k)
\end{split}
\end{equation}

which yields exactly the same form as a single-layer linear model with a softmax activation. As a result, we can use the gradient descent to train the softmax's parameters $\boldsymbol{\Theta}$, which is initialized by $\boldsymbol{M}$, in order to reduce the overall loss. Considering the convenience of derivative computation, we choose multiclass cross entropy, which is commonly used together with the softmax function, as our loss function. The overall loss on a single data point can be formulated as
\begin{equation*}
    J = -\sum_{k=1}^K{(1-y_k)log(1-{\hat{y}_k})+y_klog({\hat{y}_k})},
    \textrm{\ \ where $\boldsymbol{\Theta}$ is updated by \ }\boldsymbol{\theta_k}^t = \boldsymbol{\theta_k}^{t-1} - \gamma_1 \frac{\partial J}{\partial \boldsymbol{\theta_k}^{t-1}},
\end{equation*} 

where $\gamma_1$ is the learning rate, $\boldsymbol{y}$ is a one-hot vector transformed from the original label.
This optimization process is called by \textbf{TrainDecoding} in Alg.~\ref{algo_decoding}. Like ordinary gradient descent, data are partitioned into mini batches which are used to calculate current gradients for a single round of update. We can also apply the L1/L2 regularization here to improve the generalization ability. Note that, the validity of gradient descent guarantees the overall loss to decrease through iterations, which ensures this algorithm is a valid method to refine the decoding strategy.

\presec
\subsection{Coding Matrix Optimization}\label{codematrix_update}
\postsec

Besides decoding optimization, it is quite beneficial to optimize coding matrix through the iterative training as well. We notice that, if the input $\boldsymbol{o}$ of softmax decoding can also be updated via back propagation, we are able to further lower the overall training loss. The corresponding update process can be defined as
$\boldsymbol{o}^t = \boldsymbol{o}^{t-1} - \gamma_2 \frac{\partial J}{\partial \boldsymbol{o}^{t-1}}$, where $\gamma_2$ is the learning rate.
However, $\boldsymbol{o}$ cannot be updated directly since it is the output of base learners. Fortunately, optimizing the coding matrix $\boldsymbol{M}$ enables us to update the $\boldsymbol{o}$ indirectly so as to further reduce the overall training loss.

As stated in Sec.~\ref{sec_ecoc}, $M_{kj}$ determines the label of the data belonging to class $k$ when they are used to train base learner $j$. If we assume that base learners are able to fit the given learning target perfectly, then for any classifier $j$, its output for any data belonging to class $k$ will always satisfy $o_j^i=M_{kj}$. Thus, the changes of $\boldsymbol{M}$ will affect the targets of base learners, and then the output $\boldsymbol{o}$ of base learners will be changed subsequently. Moreover, since the gradient $\frac{\partial J}{\partial M_{kj}}$ is equal to $G_{ij}=\frac{\partial J}{\partial o_j^i}$ in this situation, we can optimize $\boldsymbol{M}$ by gradient descent: 
$M_{kj}^t = M_{kj}^{t-1} - \gamma_2 \frac{\partial J}{\partial M_{kj}^{t-1}} = M_{kj}^{t-1} - \gamma_2 \frac{\partial J}{\partial G_{ij}}$.

However, there is no perfect base learner in practice. 
As a result, we cannot use above solution to optimize $\boldsymbol{M}$ directly since $\frac{\partial J}{\partial M_{kj}} \neq G_{ij}$. Nevertheless, there are many data samples that can be used for a single class $k$. That is, for a $M_{kj}$, there are many $G_{ij}$, where $y_i = k$. So instead of using unstable gradient point $G_{ij}$, we can use average gradient of each class to have a more stable estimation for $\frac{\partial J}{\partial M_{kj}}$:
\begin{equation}\label{average_gradient}
    \frac{\partial J}{\partial M_{kj}} \vcentcolon= \frac{1}{\vert \boldsymbol{\Omega_k} \vert}\sum_{i \in \boldsymbol{\Omega_k}}{G_{ij}}, \textrm{\ \ where \ }\boldsymbol{\Omega_k} = \{i|y_i=k\}
\end{equation}

Then this estimation can be used to update the coding matrix. This optimization algorithm is described in Alg.~\ref{algo_update}, which is almost the same as a normal back propagation algorithm except using the whole batch data to calculate average gradients before performing updates. This method is also empirically proved to be effective by our experiment, as shown in the next section, which means, by optimizing global objective function, the coding matrix can be definitely refined to reduce the loss as well as enhance the generalization capability.

\presec
\subsection{Discussion}\label{lightmc_discussion}
\postsec

In the rest of section, we take further discussions about the efficiency and performance of LightMC.

\begin{itemize}[leftmargin=12pt ,itemsep=1pt,topsep=1pt]
\item \emph{Efficiency:} Compared with existing ECOC-based methods, LightMC is more efficient as it can use much less time to find a coding matrix before training. Meanwhile, it can even produce the comparable performance since the coding matrix will be dynamically refined in the subsequent training. Moreover, LightMC only requires little additional optimization computation cost, which is the same as the cost of single layer linear model and much smaller than the cost of powerful base learners like the neural networks and GBDT. The experimental results in the following section will further demonstrate the efficiency of LightMC.

\item \emph{Mini-Batch Coding Optimization Method:} One shortage of Alg.~\ref{algo_update} is inefficient in memory usage as it uses the full batch to update. Actually, it is quite natural to switch to mini-batch update since the average gradients can be calculated in mini-batches as well. 

\item \emph{Distributed Coding:} Binary coding is used in most existing ECOC-based methods. On the other hand, LightMC employs the distributed coding to perform the continuous optimization. Apparently, distributed coding, also called embedding, contains more information than binary coding~\cite{mikolov2013distributed,zhao2013sparse}, which enables LightMC to leverage more information over the correlations among classes.

\item \emph{Alternating Training with Base Learners:} As shown in Alg.~\ref{algo_train}, when the base learner is not the boosting learner, for example, the neural networks, LightMC can be called at each iteration(epoch). For the boosting learners, LightMC is conducted starting from $i_s$-th round and called once per $1/\alpha$ round. It is because there is a \textit{learning rate} $\alpha$, which will shrinkage the output of model at each iteration, in boosting learners. As a result, boosting learners need more iterations to fit the new training targets. Therefore, using initial rounds $i_s$ and being called once per $1/\alpha$ round can improve the efficiency, since calling LightMC at each iteration is not necessary.

\item \emph{Compared with Softmax Layer in Neural Networks:} The form of softmax decoding is similar to the softmax layer in neural networks. However, they are different indeed: 1) the softmax layer is actually the same to OVA decomposition, and it does not use coding matrix to encode the correlations among classes; 2) they use different optimization schemes: the loss per sample is reduced in the optimization of softmax layer, while softmax decoding optimizes the loss per class (see Eq.~\ref{average_gradient}). It is hard to say which one is better in practice for neural networks, even some recent works found the accuracy is almost the same while using fixed softmax layer~\cite{hoffer2018fix}. This topic, however, is not in the scoop of this paper.
\end{itemize}

\presec
\vspace{-5pt}
\section{Experiment}\label{experiment}
\postsec

\subsection{Experiment Setting}
\postsec

\begin{table}[t]
\centering
\precap
\vspace{-5pt}
\caption{Datasets used for experiments.}
\label{datainfo}
\begin{tabular}{lccccr}
\toprule
Dataset & \#class & \#feature & \#data\\
\midrule
News20~\cite{chang2011libsvm} & 20&    62,021&     19,928&  \\
Aloi~\cite{chang2011libsvm}&       1,000&  128&    108,000&   \\
Dmoz~\cite{yen2016pd}&       11,878& 833,484& 373,408& \\
LSHTC1~\cite{yen2016pd}&     12,045& 347,255& 83,805& \\
AmazonCat-14K~\cite{mcauley2015inferring, mcauley2015image} \footnote{\url{http://manikvarma.org/downloads/XC/XMLRepository.html}} & 14,588 \footnote{Number of class is 3344 after converting to multi-class format}&    597,940&     5,497,775&  \\
\bottomrule
\end{tabular}
\end{table}

In this section, we report the experimental results regarding our proposed LightMC algorithm. We conduct experiments on five public datasets, as listed in Table~\ref{datainfo}. From this table, we can see a wide range of the sizes of datasets, the largest of which has millions of samples with ten thousand classes and can be used to validate the scalability of LightMC. Among them, AmazonCat-14K is originally a multilabel dataset; we convert it to a multi-class one by randomly sampling one label per data. As stated in Sec.~\ref{relatedwork}, to the best of our knowledge, it is the first time to examine ECOC-based methods on such large-scale datasets. 

For the baselines, we use OVA and evolutionary ECOC proposed in~\cite{bautista2012minimal}. OVO is excluded in baselines due to its extremely inefficiency of using $K\times(K-1)/2$ base learners. For example, in \textbf{LSHTC1} data, OVO needs 72 million base learners and is estimated to take about 84 days to run an experiment even when the cost of one base learner is 0.1 second. The initial coding matrix of LightMC is set to be exactly the same as the ECOC baseline to make them comparable. Besides, to see the efficiency of LightMC, we add another LightMC baseline, but starting from random coding matrix, called LightMC(R). As for the length of the coding matrix $L$, a length of $10log_2(K)$ was suggested in~\cite{allwein2000reducing}. Considering that our base learner is more powerful, we set the $L$ to $min(5log_2(K-1)+1, K/2)$.

For all decomposition methods we use LightGBM~\cite{ke2017lightgbm} as to train base learners. In all experiments we set \textit{learning\_rate} ($\alpha$) to $0.1$, \textit{num\_leaves} (max number of leaves in a single tree) to $127$ and \textit{early\_stopping} (early stopping rounds) to 20. For the AmazonCat-14K, we override \textit{num\_leaves} to 300 and \textit{early\_stopping} to 10, otherwise it needs several weeks to run a experiment. Other parameters remain to be the same as default. Our experimental environment is a Windows server with two E5-2670 v2 CPUs (in total 20 cores) and 256GB memories. All experiments run with multi-threading and the number of threads is fixed to 20. 

Regarding parameters used by LightMC, the starting round $i_s$ is set to $30$, $\gamma_1$ to $0.1$ and $\gamma_2$ to $0.2$. And softmax's parameters $\Theta$ are trained for one epoch each time the optimization method is called. 

\presec
\vspace{-10pt}
\subsection{Experiment Result Analysis}
\postsec

\begin{table}[h]
    \centering
    \precap
    \caption{Comparison on test classification error, lower is better.}
    \label{error_result}
    \begin{tabular}{lrrrrrrr}
    \toprule
    Dataset &   OVA & ECOC & LightMC(R) & LightMC\\
    \midrule
    News20& 18.66\%   & 20.82\% $\pm$ 0.33\%& 20.63\% $\pm$ 0.57\%& \textbf{18.63\% $\pm$ 0.37\%}\\
    Aloi&   11.44\%   & 10.72\% $\pm$ 0.12\%& 10.75\% $\pm$ 0.23\%& \textbf{9.75\% $\pm$ 0.12\%}\\
    Domz&     N/A     & 55.87\% $\pm$ 0.34\%& 55.55\% $\pm$ 0.44\%& \textbf{53.95\% $\pm$ 0.25\%}\\
    LSHTC1&   N/A     & 76.04\% $\pm$ 0.59\%& 76.17\% $\pm$ 0.73\%& \textbf{75.63\% $\pm$ 0.33\%}\\
    AmazonCat-14K& N/A& 27.05\% $\pm$ 0.11\%& 26.98\% $\pm$ 0.21\%& \textbf{25.54\% $\pm$ 0.10\%}\\
    \bottomrule
    \end{tabular}
\end{table}

\begin{table}[hb]
    \centering
    \precap
    \caption{Training convergence time (average of multiple runs) and coding matrix searching time (the last column) in second.}
    \label{timing_result}
    \begin{tabular}{lrrrr|rr}
    \toprule
    Dataset &   OVA  & ECOC & LightMC(R) & LightMC & \textit{Coding Matrix}\\
    \midrule
    News20& \textbf{71}            & 120     & 133    & 100     & \textit{34}      \\
    Aloi&   1494          & 717     & 753    & \textbf{627}     & \textit{201}     \\
    Domz&   > 259k        & 58,320  & 61,930 & \textbf{51,840}  & \textit{13,233}  \\
    LSHTC1& > 86k         & 5,796   & 5,995  & \textbf{5,690}   & \textit{926}    \\
    AmazonCat-14K& > 969k & 332,280 & 354,480& \textbf{311,040} & \textit{48,715}  \\
    \bottomrule
    \end{tabular}
\end{table}

\begin{table}[t]
    \centering
    \setlength\tabcolsep{3pt}
    \precap
    \caption{Code distances in LightMC over class pairs on \textbf{News20} dataset, regarding iterations.}
    \label{dist_result}
    \begin{tabular}{lrrrrrrrrrrrrrrr}
    \toprule
    Class Pairs    &   0  & 50 & 100 & 150 & 200 & 300 & 400 & 500 & 1000 \\
    \midrule
    ibm.hardware, mac.hardware     &98.9  & 97.4   & 96.8    & 95.9     & 95.1  & 93.1 & 91.1 & 89.3 &81.8   \\
    mac.hardware, politics.mideast &120.9 & 135.5  & 136.4   & 136.8    & 140.8 & 145.6  & 149.5 & 152.7  &163.2\\
    \bottomrule
    \end{tabular}
\end{table}

\begin{figure}[ht]
    \begin{subfigure}{0.48\textwidth}
        \centering
        \includegraphics[width=6.7cm, height=4.5cm]{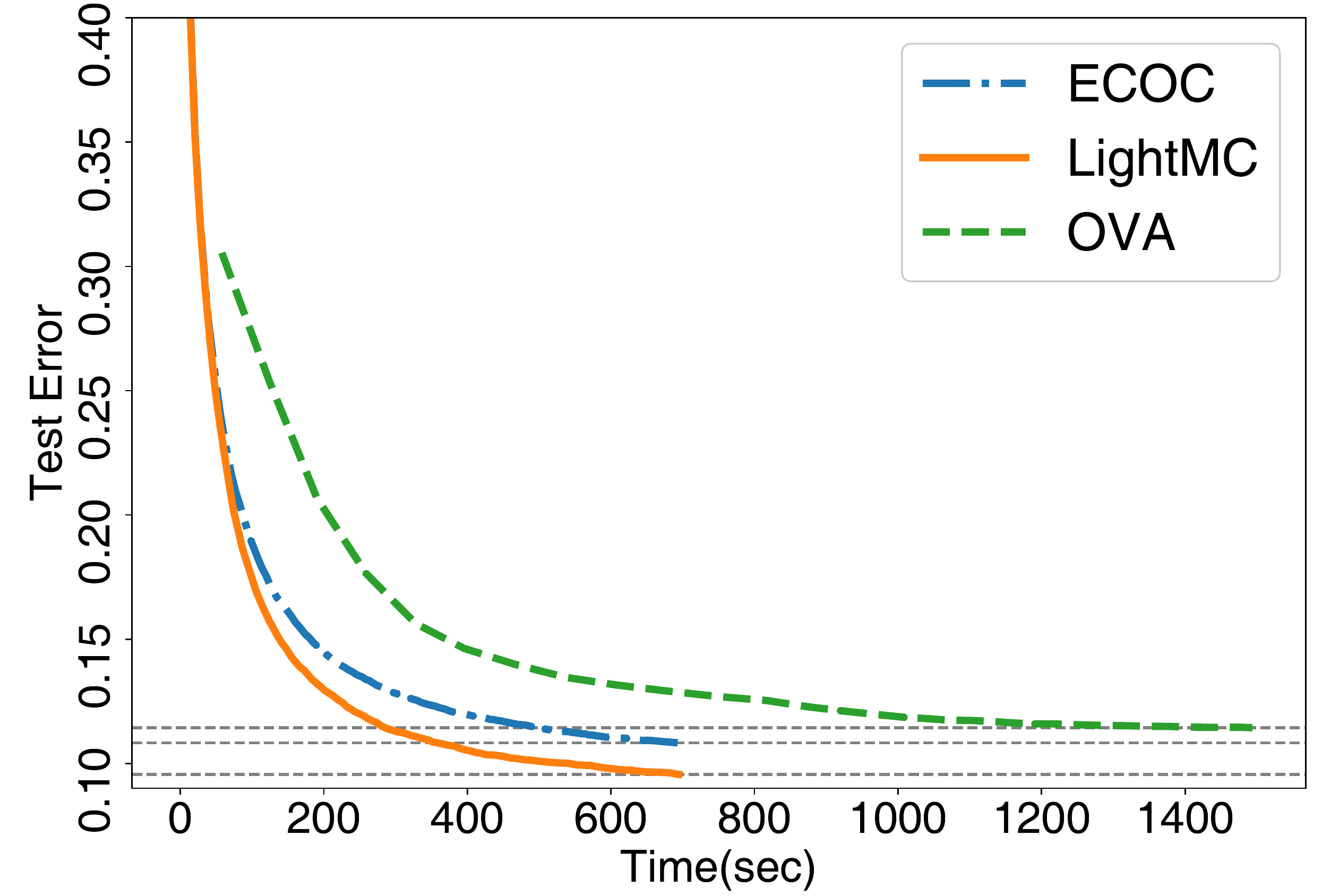}
        \caption{Aloi}\label{aloi_performance}
        
    \end{subfigure}
    \hspace{2mm}
    \begin{subfigure}{0.48\textwidth}
        \centering
        \includegraphics[width=6.7cm, height=4.5cm]{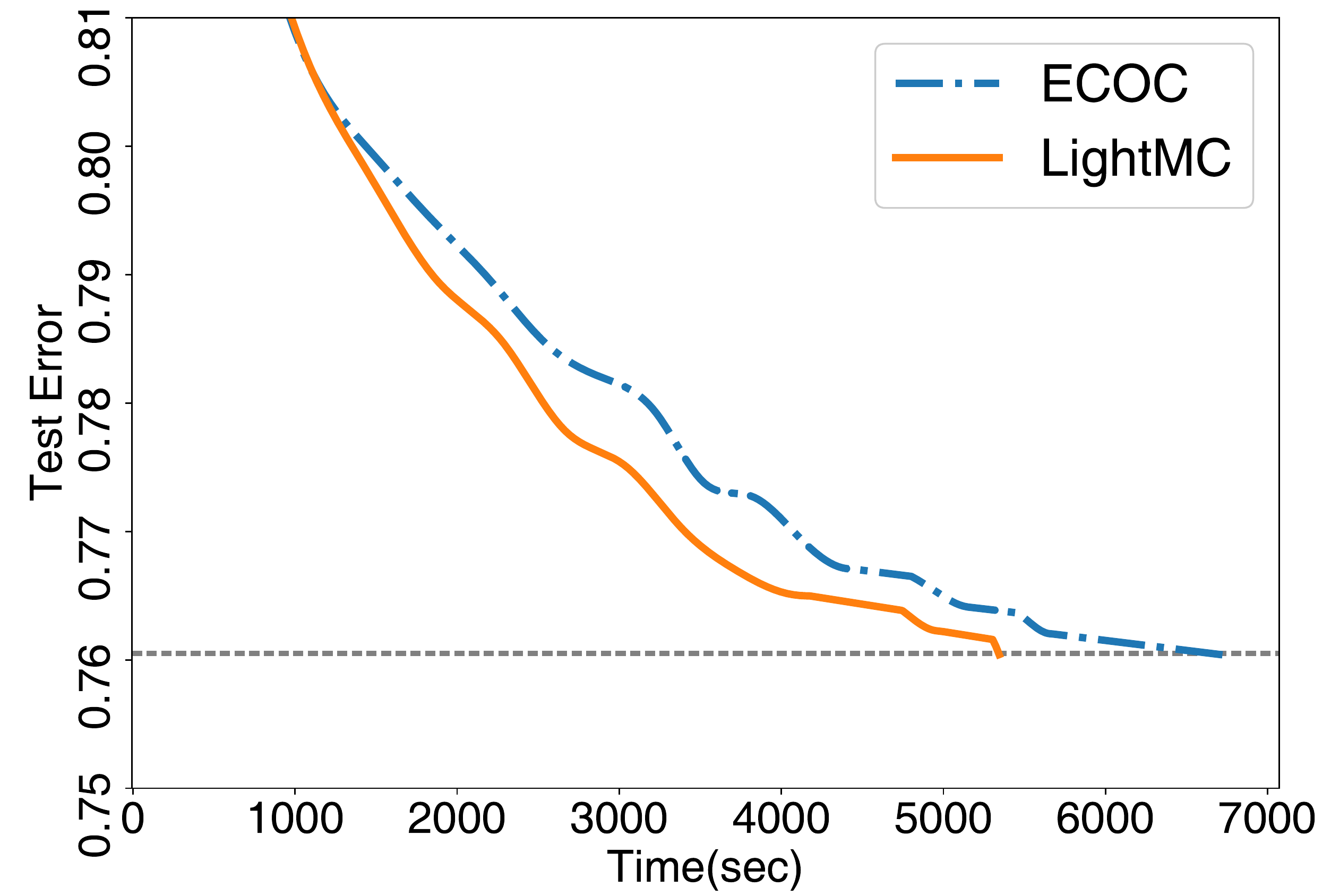}
        \caption{LSHTC1}\label{LSHTC1_performance}
        \vspace{-5pt}
    \end{subfigure}
    \caption{Convergence curves on \textbf{Aloi} and \textbf{LSHTC1} datasets.}
    \postcap
\end{figure}

The experiment results are reported in Table~\ref{error_result} and~\ref{timing_result}. The OVA error result on \textbf{Dmoz}, \textbf{LSHTC1} and \textbf{Amazon-Cat-14k} datasets are not reported since the time costs are extremely too high. However, we estimate their convergence time by using the subset of the original data.

From these two tables, we find LightMC outperforms all the others in terms of both accuracy and convergence time. In particular, both ECOC and LightMC yield faster convergence over OVA when $K$ is larger. Furthermore, compared with ECOC, LightMC increases the accuracy by about 3\% (relatively), and improves 5.88\% at the best case on the \textbf{LSHTC1} dataset. As for the speed, LightMC also uses less time than ECOC to converge. These results clearly indicate that LightMC can further reduce the overall loss by dynamically refining the coding and decoding strategy as expected.

We can also find that, while starting from random coding matrix, the accuracy of LightMC(R) is comparable with that of ECOC. Despite the slower convergence of LightMC(R), the total time cost of LightMC(R) is still much less than ECOC, since ECOC spends an enormous additional time to find a good coding matrix before training. This result further implies the efficiency of LightMC: it can provide comparable accuracy without searching a sub-optimal coding matrix before training. 

To demonstrate more learning details, we plot the curves of the test error regarding the training time on \textbf{Aloi} and \textbf{LSHTC1} datasets, as shown in Fig.~\ref{aloi_performance} and~\ref{LSHTC1_performance}, respectively. From Fig.~\ref{aloi_performance}, we can see clearly that the curve of LightMC always stays below the curves of the other two methods and converges earliest at the lowest point. Fig.~\ref{LSHTC1_performance} shows a slightly different pattern: LightMC and ECOC have similar accuracy and take comparable time to converge. However, LightMC still always stays below ECOC and converges 1,405 seconds earlier than ECOC, which also indicates that LightMC succeeds in enhancing existing ECOC methods.

In addition, to illustrate the effects of LightMC in optimizing the code matrix, we calculate the distances of some class pairs, on \textbf{News20}, over the optimized coding matrix. As shown in Table~\ref{dist_result}, the distance over the class pair (`{\em ibm.hardware}',`{\em mac.hardware}') is obviously much smaller than that over (`{\em mac.hardware}',`{\em politics.mideast}'). Moreover, the distance over the former pair keeps reducing along with the training of LightMC, while that over the latter, on the other hand, keeps increasing due to the irrelevance between this class pair. This result empirically implies the effectiveness of LightMC in optimizing the coding matrix towards to the right direction.

As a summary, all these results have illustrated the effectiveness and efficiency of LightMC. LightMC cannot only empower existing ECOC-based methods but also achieve the comparable classification accuracy using much less time since it saves the time for finding a sound coding matrix. Moreover, LightMC can optimize the coding matrix towards to the better direction.

\presec
\section{Conclusion}\label{conclusion}
\vspace{-10pt}

We propose a novel dynamic ECOC-based multiclass decomposition algorithm, named LightMC, to solve large-scale classification problems efficiently. To leverage better of correlations among classes, LightMC dynamically optimizes its coding matrix and decoding strategy, jointly with the training of base learners. Specifically, we design a new differentiable decoding strategy to enable direct optimization over the decoding strategy and coding matrix. Experiments on public datasets with classes ranging from twenty to more than ten thousand empirically show the effectiveness and the efficiency of LightMC. In future, we plan to examine how LightMC will work while replacing the softmax layer in neural networks.

\newpage
\bibliographystyle{unsrt}
\bibliography{Main}

\begin{thebibliography}{10}

\bibitem{ciregan2012multi}
Dan Ciregan, Ueli Meier, and J{\"u}rgen Schmidhuber.
\newblock Multi-column deep neural networks for image classification.
\newblock In {\em Computer vision and pattern recognition (CVPR), 2012 IEEE
  conference on}, pages 3642--3649. IEEE, 2012.

\bibitem{nigam2000text}
Kamal Nigam, Andrew~Kachites McCallum, Sebastian Thrun, and Tom Mitchell.
\newblock Text classification from labeled and unlabeled documents using em.
\newblock {\em Machine learning}, 39(2-3):103--134, 2000.

\bibitem{schulten2001commerce}
Ellen Schulten, Hans Akkermans, Guy Botquin, Martin D{\"o}rr, Nicola Guarino,
  Nelson Lopes, and Norman Sadeh.
\newblock The e-commerce product classification challenge.
\newblock {\em IEEE Intelligent systems}, 16(4):86--89, 2001.

\bibitem{panca2017application}
V~Panca and Z~Rustam.
\newblock Application of machine learning on brain cancer multiclass
  classification.
\newblock In {\em AIP Conference Proceedings}, volume 1862, page 030133. AIP
  Publishing, 2017.

\bibitem{bredensteiner1999multicategory}
Erin~J Bredensteiner and Kristin~P Bennett.
\newblock Multicategory classification by support vector machines.
\newblock In {\em Computational Optimization}, pages 53--79. Springer, 1999.

\bibitem{choromanska2015logarithmic}
Anna~E Choromanska and John Langford.
\newblock Logarithmic time online multiclass prediction.
\newblock In {\em Advances in Neural Information Processing Systems}, pages
  55--63, 2015.

\bibitem{mroueh2012multiclass}
Youssef Mroueh, Tomaso Poggio, Lorenzo Rosasco, and Jean-Jeacques Slotine.
\newblock Multiclass learning with simplex coding.
\newblock In {\em Advances in Neural Information Processing Systems}, pages
  2789--2797, 2012.

\bibitem{weston1998multi}
Jason Weston and Chris Watkins.
\newblock Multi-class support vector machines.
\newblock Technical report, Citeseer, 1998.

\bibitem{hsu2009multi}
Daniel~J Hsu, Sham~M Kakade, John Langford, and Tong Zhang.
\newblock Multi-label prediction via compressed sensing.
\newblock In {\em Advances in neural information processing systems}, pages
  772--780, 2009.

\bibitem{prabhu2014fastxml}
Yashoteja Prabhu and Manik Varma.
\newblock Fastxml: A fast, accurate and stable tree-classifier for extreme
  multi-label learning.
\newblock In {\em Proceedings of the 20th ACM SIGKDD international conference
  on Knowledge discovery and data mining}, pages 263--272. ACM, 2014.

\bibitem{si2017gradient}
Si~Si, Huan Zhang, S~Sathiya Keerthi, Dhruv Mahajan, Inderjit~S Dhillon, and
  Cho-Jui Hsieh.
\newblock Gradient boosted decision trees for high dimensional sparse output.
\newblock In {\em International Conference on Machine Learning}, pages
  3182--3190, 2017.

\bibitem{yen2016pd}
Ian En-Hsu Yen, Xiangru Huang, Pradeep Ravikumar, Kai Zhong, and Inderjit
  Dhillon.
\newblock Pd-sparse: A primal and dual sparse approach to extreme multiclass
  and multilabel classification.
\newblock In {\em International Conference on Machine Learning}, pages
  3069--3077, 2016.

\bibitem{nilsson1965learning}
Nils~J Nilsson, Terrence~J Sejnowski, and Halbert White.
\newblock Learning machines.
\newblock 1965.

\bibitem{hastie1998classification}
Trevor Hastie and Robert Tibshirani.
\newblock Classification by pairwise coupling.
\newblock In {\em Advances in neural information processing systems}, pages
  507--513, 1998.

\bibitem{dietterich1995solving}
Thomas~G Dietterich and Ghulum Bakiri.
\newblock Solving multiclass learning problems via error-correcting output
  codes.
\newblock {\em Journal of artificial intelligence research}, 2:263--286, 1995.

\bibitem{crammer2002learnability}
Koby Crammer and Yoram Singer.
\newblock On the learnability and design of output codes for multiclass
  problems.
\newblock {\em Machine learning}, 47(2-3):201--233, 2002.

\bibitem{baro2009traffic}
Xavier Bar{\'o}, Sergio Escalera, Jordi Vitri{\`a}, Oriol Pujol, and Petia
  Radeva.
\newblock Traffic sign recognition using evolutionary adaboost detection and
  forest-ecoc classification.
\newblock {\em IEEE Transactions on Intelligent Transportation Systems},
  10(1):113--126, 2009.

\bibitem{pujol2006discriminant}
Oriol Pujol, Petia Radeva, and Jordi Vitria.
\newblock Discriminant ecoc: A heuristic method for application dependent
  design of error correcting output codes.
\newblock {\em IEEE Transactions on Pattern Analysis and Machine Intelligence},
  28(6):1007--1012, 2006.

\bibitem{garcia2008evolving}
Nicolas Garcia-Pedrajas and Colin Fyfe.
\newblock Evolving output codes for multiclass problems.
\newblock {\em IEEE Transactions on Evolutionary Computation}, 12(1):93--106,
  2008.

\bibitem{bautista2012minimal}
Miguel~{\'A}ngel Bautista, Sergio Escalera, Xavier Bar{\'o}, Petia Radeva,
  Jordi Vitri{\'a}, and Oriol Pujol.
\newblock Minimal design of error-correcting output codes.
\newblock {\em Pattern Recognition Letters}, 33(6):693--702, 2012.

\bibitem{bagheri2013genetic}
Mohammad~Ali Bagheri, Qigang Gao, and Sergio Escalera.
\newblock A genetic-based subspace analysis method for improving
  error-correcting output coding.
\newblock {\em Pattern Recognition}, 46(10):2830--2839, 2013.

\bibitem{bautista2014design}
Miguel~{\'A}ngel Bautista, Sergio Escalera, Xavier Bar{\'o}, and Oriol Pujol.
\newblock On the design of an ecoc-compliant genetic algorithm.
\newblock {\em Pattern Recognition}, 47(2):865--884, 2014.

\bibitem{zhang2009spectral}
Xiao Zhang, Lin Liang, and Heung-Yeung Shum.
\newblock Spectral error correcting output codes for efficient multiclass
  recognition.
\newblock In {\em Computer Vision, 2009 IEEE 12th International Conference on},
  pages 1111--1118. IEEE, 2009.

\bibitem{zhao2013sparse}
Bin Zhao and Eric~P Xing.
\newblock Sparse output coding for large-scale visual recognition.
\newblock In {\em Computer Vision and Pattern Recognition (CVPR), 2013 IEEE
  Conference on}, pages 3350--3357. IEEE, 2013.

\bibitem{hatami2012thinned}
Nima Hatami.
\newblock Thinned-ecoc ensemble based on sequential code shrinking.
\newblock {\em Expert Systems with Applications}, 39(1):936--947, 2012.

\bibitem{rocha2014multiclass}
Anderson Rocha and Siome~Klein Goldenstein.
\newblock Multiclass from binary: Expanding one-versus-all, one-versus-one and
  ecoc-based approaches.
\newblock {\em IEEE Transactions on Neural Networks and Learning Systems},
  25(2):289--302, 2014.

\bibitem{escalera2006ecoc}
Sergio Escalera and Oriol Pujol.
\newblock Ecoc-one: A novel coding and decoding strategy.
\newblock In {\em Pattern Recognition, 2006. ICPR 2006. 18th International
  Conference on}, volume~3, pages 578--581. IEEE, 2006.

\bibitem{escalera2010decoding}
Sergio Escalera, Oriol Pujol, and Petia Radeva.
\newblock On the decoding process in ternary error-correcting output codes.
\newblock {\em IEEE transactions on pattern analysis and machine intelligence},
  32(1):120--134, 2010.

\bibitem{mikolov2013distributed}
Tomas Mikolov, Ilya Sutskever, Kai Chen, Greg~S Corrado, and Jeff Dean.
\newblock Distributed representations of words and phrases and their
  compositionality.
\newblock In {\em Advances in neural information processing systems}, pages
  3111--3119, 2013.

\bibitem{hoffer2018fix}
Elad Hoffer, Itay Hubara, and Daniel Soudry.
\newblock Fix your classifier: the marginal value of training the last weight
  layer.
\newblock {\em arXiv preprint arXiv:1801.04540}, 2018.

\bibitem{chang2011libsvm}
Chih-Chung Chang and Chih-Jen Lin.
\newblock Libsvm: a library for support vector machines.
\newblock {\em ACM transactions on intelligent systems and technology (TIST)},
  2(3):27, 2011.

\bibitem{mcauley2015inferring}
Julian McAuley, Rahul Pandey, and Jure Leskovec.
\newblock Inferring networks of substitutable and complementary products.
\newblock In {\em Proceedings of the 21th ACM SIGKDD International Conference
  on Knowledge Discovery and Data Mining}, pages 785--794. ACM, 2015.

\bibitem{mcauley2015image}
Julian McAuley, Christopher Targett, Qinfeng Shi, and Anton Van Den~Hengel.
\newblock Image-based recommendations on styles and substitutes.
\newblock In {\em Proceedings of the 38th International ACM SIGIR Conference on
  Research and Development in Information Retrieval}, pages 43--52. ACM, 2015.

\bibitem{allwein2000reducing}
Erin~L Allwein, Robert~E Schapire, and Yoram Singer.
\newblock Reducing multiclass to binary: A unifying approach for margin
  classifiers.
\newblock {\em Journal of machine learning research}, 1(Dec):113--141, 2000.

\bibitem{ke2017lightgbm}
Guolin Ke, Qi~Meng, Thomas Finley, Taifeng Wang, Wei Chen, Weidong Ma, Qiwei
  Ye, and Tie-Yan Liu.
\newblock Lightgbm: A highly efficient gradient boosting decision tree.
\newblock In {\em Advances in Neural Information Processing Systems}, pages
  3149--3157, 2017.

\end{thebibliography}

\end{document}